\pdfoutput=1

\documentclass[11pt]{article}

\usepackage{EMNLP2022}
\usepackage{graphicx}
\usepackage{times}
\usepackage{latexsym}
\usepackage{times}
\usepackage{latexsym}
\usepackage{amssymb}
\usepackage{booktabs}
\usepackage{amsfonts}
\usepackage{caption}
\usepackage{threeparttable}
\usepackage{geometry}
\usepackage{multirow}
\usepackage[T1]{fontenc}

\usepackage[utf8]{inputenc}

\usepackage{microtype}

\usepackage{inconsolata}
\usepackage{multirow, makecell}
\usepackage{rotating}
\usepackage{mathtools}
\usepackage[T1]{fontenc}
\usepackage{colortbl}
\newcommand*{\affmark}[1][*]{\textsuperscript{#1}}
\makeatletter
\def\thanks#1{\protected@xdef\@thanks{\@thanks
        \protect\footnotetext{#1}}}
\makeatother

\usepackage[utf8]{inputenc}
\usepackage{amsmath}
\usepackage{graphicx}
\usepackage{microtype}
\usepackage{booktabs}

\definecolor{gred}{RGB}{255,102,102}
\definecolor{gblue}{RGB}{51,102,255}
\definecolor{gyellow}{RGB}{244,180,0}
\definecolor{ggreen}{RGB}{15,157,88}
\definecolor{ggrey}{RGB}{115,115,115}
\definecolor{na}{gray}{0.9}
\definecolor{LightYellow}{RGB}{255,255,191}
\definecolor{OrangeRed}{rgb}{1.0, 0.27, 0.0}
\definecolor{midnightgreen}{rgb}{0.0, 0.29, 0.33}
\definecolor{darkgreen}{rgb}{0.0, 0.42, 0.24}

\newcommand{\colorG}[1]{\textcolor{ggreen}{\textbf{#1}}}
\newcommand{\colorB}[1]{\textcolor{gblue}{\textbf{#1}}}

\usepackage{threeparttable, tablefootnote}
\definecolor{skyblue}{RGB}{70, 130, 180}
\usepackage{CJKutf8}

\title{Revisiting DocRED -- Addressing the False Negative Problem \\ in Relation Extraction}

\author{Qingyu Tan\affmark[* 1, 2]
\thanks{$^*$ Equal contribution.  Qingyu Tan and Lu Xu are under the Joint PhD Program between Alibaba and NUS/SUTD. }~~~\textbf{Lu Xu\affmark[* 1, 3]~~~Lidong Bing\affmark[$^\dag$ 1] \thanks{$^\dag$  Corresponding author.}~~~Hwee Tou Ng\affmark[2]~~~Sharifah Mahani Aljunied\affmark[1]} 
\\$^1$DAMO Academy, Alibaba Group~~\\
$^2$Department of Computer Science, National University of Singapore\\
$^3$Singapore University of Technology and Design \\
\texttt{\{qingyu.tan,lu.x,l.bing,mahani.aljunied\}@alibaba-inc.com}\\
\texttt{\{qtan6,nght\}@comp.nus.edu.sg}\\
}

\begin{document}
\maketitle
\begin{abstract}
The DocRED dataset is one of the most popular and widely used benchmarks for document-level relation extraction (RE). 
It adopts a recommend-revise annotation scheme so as to have  a large-scale annotated dataset.
However, we find that the annotation of DocRED is incomplete, i.e., false negative samples are prevalent. 
We analyze the causes and effects of the overwhelming false negative problem in the DocRED dataset. 
To address the shortcoming, we re-annotate 4,053 documents in the DocRED dataset by adding the missed relation triples back to the original DocRED. 
We name our revised DocRED dataset Re-DocRED. 
We conduct extensive experiments with state-of-the-art neural models on both datasets, and the experimental results show that the models trained and evaluated on our Re-DocRED achieve performance improvements of around 13 F1 points. 
Moreover, we conduct a comprehensive analysis to identify the potential areas for further improvement.\footnote{Our dataset is publicly available at \url{https://github.com/tonytan48/Re-DocRED}.}
\end{abstract}

\section{Introduction}
The field of relation extraction (RE) is related to knowledge bases (KBs). Most popular relation extraction datasets are constructed from knowledge triples in KBs. For example, the TACRED dataset~\citep{zhang-etal-2017-position} is constructed by the TAC Knowledge Base Population challenge. The NYT10 \citep{riedel-etal-2013-relation} dataset matched the Freebase Knowledge base \citep{bollacker2008freebase} to the New York Times corpus \citep{sandhaus2008new}. Wiki20~\citep{gao2021manual} and DocRED~\citep{yao2019docred} originated from distant supervision from the Wikidata knowledge base~\citep{vrandevcic2014wikidata} and Wikipedia articles. 
By exploiting distant supervision, relation triple candidates can be retrieved from the knowledge bases for a given piece of text. 
The retrieved triples are based on entity co-occurrence only, and they may not be related to the context. This is known as the false positive (FP) problem (\citealp{bing-etal-2015-improving}; \citealp{xiao-etal-2020-denoising}). 
\begin{figure}[!t]
\centering{
\small{
\begin{tabular}{|p{0.92\columnwidth}|}
\hline
\vspace{-.1cm}
"\textbf{I Knew You Were Trouble}~" is a song recorded by American singer - songwriter \textbf{Taylor Swift} for her fourth studio album , \textbf{Red} ( 2012 ) . It was released on \textbf{October 9 , 2012} , in \textbf{the United States} by \textbf{Big Machine Records} as the third promotional single from the album . Later , " \textbf{I Knew You Were Trouble} " was released as the third single from \textbf{Red} on \textbf{November 27 , 2012} , in \textbf{the United States }. It was written by \textbf{Swift} , \textbf{Max Martin} and \textbf{Shellback} , \colorbox{LightYellow}{\underline{with the production handled by the latter two}} ... It later peaked at number two in \textbf{January 2013} , blocked from the top spot by \textbf{Bruno Mars}’ " \textbf{Locked Out of Heaven} " . At the inaugural \textbf{YouTube Music Awards }in 2013 , " \textbf{I Knew You Were Trouble }" won the award for \textbf{YouTube} phenomenon …  \\
\vspace{0.05cm}
\textcolor{orange}{\textbf{DocRED}}: (\textbf{I Knew You Were Trouble}, \textit{producer}, \textbf{Max Martin}); (\textbf{Taylor Swift},~\textit{country of citizenship},~ \textbf{the United States}) ...  \\
\vspace{0.05cm}
\textcolor{skyblue}{\textbf{Re-DocRED}}:  (\textbf{I Knew You Were Trouble}, \textit{producer}, \textbf{Max Martin}); (\textbf{I Knew You Were Trouble},~\textit{producer},~\textbf{Shellback}) ... 
\vspace{.1cm}
\\
\hline
\end{tabular}
}}
\caption{A sample document in our Re-DocRED dataset. From the \colorbox{LightYellow}{\underline{highlighted evidence}}, it can be inferred that ``I Knew You Were Trouble'' was produced by both ``Max Martin'' and ``Shellback'', but in the previous incomplete DocRED dataset, only the first producer is labeled.}
\label{fig:example-doc}
\end{figure}


However, the false negative (FN) problem in the relation extraction datasets receives less attention as compared to FPs. Without resolving this issue, the annotations of the datasets are incomplete.
Recent efforts on addressing the false negative problem are from the model perspective (\citealp{chen2021h}; \citealp{hao-etal-2021-knowing}), which aim to denoise the false negative data during training.
The challenge for these approaches is that both development and test sets can be incomplete at the same time. Without a completely annotated dataset, the current evaluation is ill-defined.


Several re-annotation works revised the existing sentence-level relation extraction datasets. 
\citet{alt2020tacred} re-annotated a small amount of challenging samples in the development and test sets of the TACRED dataset \citep{zhang-etal-2017-position}. \citet{stoica2021re} extended this work and re-annotated the training, development, and test sets of TACRED with a semantically refined label space. Besides, \citet{gao2021manual} re-annotated the test sets of two distantly supervised RE datasets. 
Even though their discussions and examples emphasized the efforts on correcting the false positive samples, all
the revised versions of the datasets have a significant increase of positive samples. 
That is, many samples that are previously labeled as \textit{no\_relation}~(NA) are re-annotated with relation labels in the revised datasets.
We show the detailed statistics in Table~\ref{tab:data-stats}, which indicates that the false negative problem is prevalent in the sentence-level relation extraction field. 
Compared to the sentence-level task, document-level RE is more susceptible to the false negative problem. This is primarily because document-level RE involves significantly more entity pairs in a raw text, as shown in Table \ref{tab:data-stats}. 
Note that the objective of the RE task is to determine the relation types for all entity pairs, and the number of entity pairs is quadratic in the number of entities.

In this paper, we address the false negative problem in DocRED. We find that the false negative problem arises due to two reasons. 
First, although the Wikidata knowledge base provides a good starting point for distantly supervised annotation of DocRED, it is highly sparse and far from complete. There are many relation triples that are 
not in the Wikidata KB. 
For example, in Figure~\ref{fig:example-doc}, the article reflects that the song ``I Knew You Were Trouble'' was produced by both ``Max Martin'' and ``Shellback'', but the fact that ``Shellback'' is the other producer is not included in both the knowledge base and the DocRED dataset. 
Second, the additional relation triples recommended by the RE model and human annotators cannot cover the ground-truth relation triples that are missed from the Wikidata. A detailed discussion is given in Section \ref{sec:preliminaries}. 
With the incompletely annotated development and test sets, the previous evaluation does not necessarily provide a fair reference.
Therefore, we propose to revise the DocRED dataset to recover the incomplete annotations through an iterative approach with human in the loop.
Specifically, we use multiple state-of-the-art document-level RE models to generate relation candidates and ask human annotators to examine the recommended triples. The details of our iterative human-in-the-loop annotation process are given in Section \ref{sec:revision}.

Most recently, \citet{Huang2022DoesRP} also identify the false negative issue in DocRED \cite{yao2019docred}. They combat the problem by annotating 96 documents from scratch with two expert annotators. 
However, annotating relation triples from scratch is different from revising recommended triples, and it is difficult to scale up to a dataset of a larger size.
We provide a comprehensive analysis between our approach and the annotating-from-scratch approach in Appendix \ref{app:comparisonwithscratch}.
Compared to \citet{Huang2022DoesRP}, our approach is better in the following aspects. First, our dataset is significantly larger in size (4,053 vs 96 documents). 
Second, the precision of our annotation is higher. 
Third, our evaluation dataset contains more triples per document than \citet{Huang2022DoesRP}, indicating that our dataset better tackles the incompleteness problem of DocRED. 
Fourth, our dataset annotation approach is more scalable and can be extended to an arbitrary number of relation types. 

Our contributions can be summarized as follows:
\begin{itemize}
    \item We identify the overwhelming false negative problem in relation extraction. 
    \item We show that the false negative problem is the cause of performance bottleneck on many relation extraction datasets, and we also provide a high-quality revised version of the document-level relation extraction dataset Re-DocRED. 
    \item Moreover, we conduct comprehensive experimental analysis to identify the potential areas for further improvement.
\end{itemize}

\begin{table*}[ht]
\centering
\resizebox{\textwidth}{!}{
\begin{threeparttable}
\begin{tabular}{lccccccl} 
\toprule
\textbf{Dataset}            & \multicolumn{1}{c}{\textbf{\# Relations}} & \multicolumn{1}{c}{\textbf{\# Triples}} & \multicolumn{1}{c}{\textbf{\# Samples}} & \multicolumn{1}{c}{\textbf{Avg. \# Words}} & \multicolumn{1}{c}{\textbf{Avg. \# Entities}} & \multicolumn{1}{c}{\textbf{Avg. \# Entity Pairs}} & \multicolumn{1}{c}{\textbf{NA}}  
\\ 
\midrule
\multicolumn{2}{l}{\textbf{Sentence-level}} \vspace{0.05cm} \\ 
Wiki20~\citep{han-etal-2020-data}     & 454                             & 15,026                        & 28,897                       & 25.7                             & 2                                   & 1                                       & 48.0\%                 \\
Wiki20m~\citep{gao2021manual} & 81                              & 103,709                       & 137,986                      & 25.6                             & 2                                   & 1                                       & 24.8\%{\color{green}$\downarrow$}                 \\
TACRED~\citep{zhang-etal-2017-position}     & 42                              & 21,373                        & 106,264                      & 36.4                             & 2                                   & 1                                       & 79.9\%                 \\
Re-TACRED~\citep{stoica2021re}  & 40                              & 33,690                        & 91,467                       & 36.4                             & 2                                   & 1                                       & 63.2\%{\color{green}$\downarrow$} \\ 
\midrule
\multicolumn{2}{l}{\textbf{Document-level}} \vspace{0.05cm} \\
HacRED~\citep{cheng-etal-2021-hacred}     & 26                              & 56,798                        & 7,731\tnote{1}                      & 122.6                            & 10.7                                & 163.9                                   & 95.7\%                 \\
DocRED~\citep{yao2019docred}     & 96                              & 50,503                        & 4,053\tnote{1}                        & 198.4                            & 19.5                                & 393.6                                  & 97.1\%                 \\
\textbf{Re-DocRED} (Ours)  & 96                              & 120,664                       & 4,053                        & 198.4                            & 19.4\tnote{2}                            & 391.0                                  & 94.0\%{\color{green}$\downarrow$}                  \\
\bottomrule
\end{tabular}
  \begin{tablenotes}[para]
    \item[1] We do not include the blind test sets for these two datasets.  \item[2] We also resolve the coreferential errors. Details are given in Appendix~\ref{app:coreference}.
  \end{tablenotes}
\end{threeparttable}
}
\caption{Statistics of relation extraction datasets. We can see that all re-annotation efforts on sentence-level relation extraction datasets (Wiki20 $\rightarrow$ Wiki20m, TACRED $\rightarrow$ Re-TACRED) result in significantly smaller percentages of \textit{no\_relation} (NA) samples, indicating that false negative is a common problem in the RE datasets.}
\label{tab:data-stats}
\end{table*}
\section{Preliminaries}
\label{sec:preliminaries}
\subsection{Background of the DocRED Dataset}
\label{sec:background-docred}
The DocRED dataset~\citep{yao2019docred} is one of the popular and widely-studied benchmark datasets for document-level relation extraction. 
The dataset contains 5,053 Wikipedia documents, where each document has an average length of 196.7 words, and an average of 19.5 entities.
With 97 predefined relation types (including \textit{no\_relation}) and an average of 393.6 entity pairs, there exist around 38,000 relation triple candidates per document to be annotated.
To reduce the annotation workload, a recommend-revise scheme is adopted.
Specifically, relation triple candidates are recommended from distantly supervised data and predictions of an RE model.
On average for each document, 19.9 triples are suggested from distantly supervised data and 7.8 triples are recommended by relation extraction models. Human annotators are asked to read the documents and review the relation triple candidates, and wrong recommendations will be discarded. As a result, an average of 12.5 triples per document are kept in the DocRED dataset.

\subsection{Problem of Incomplete Annotation}
\label{sec:problem-docred}
Based on our empirical analysis of the top-performing models \cite{tan2022document, zhang2021document, zhou2021document} on the DocRED leaderboard, many predicted triples are correct but are not annotated in the DocRED dataset. Therefore, it is important to review the dataset and identify the true bottleneck of document-level RE. We identify incomplete annotation as the major issue in the DocRED dataset. As mentioned in Section~\ref{sec:background-docred}, it is difficult to annotate a document-level RE dataset from scratch. The DocRED dataset is mainly created by human filtering the recommended relation triples in distantly supervised data and predictions from an RE model.
It is worth noting that the construction procedure of DocRED relies on the underlying assumption that the combination of the recommended triples by distant supervision and the RE model contains almost all the ground-truth relation triples in the documents. 
This assumption is not true since the relations in the Wikidata knowledge base (KB) are sparse and incomplete. That is, many semantically related triples are not reflected in the KB. Therefore, treating such triples as negative instances (\textit{no\_relation}) will introduce false negative examples into the dataset. 
In the original DocRED production process, only the distantly supervised data is used to train the RE model, which may lead to low-quality prediction. 
Furthermore, the performance of the previously used RE model is significantly worse than the recent approaches based on pre-trained language models.
Therefore, the recommended relation triples from the RE model are likely not enough for covering the ground-truth relation triples for a given text.

By relying heavily on the relation triple candidates generated from the above two methods, the annotation of the previous DocRED dataset is incomplete.
While the recommend-revise scheme is the major source of the incompleteness of the original DocRED dataset \citep{yao2019docred}, a secondary source of incompleteness comes from logical inconsistency. 
There are many inverse relation pairs in DocRED. 
For example, if entity A is annotated as a \textit{sibling} of entity B, then entity B is also a \textit{sibling} of entity A. The lack of inverse relation also contributes to the incompleteness problem.



\subsection{The Impact of False Negatives}

To identify the difficulty of relation extraction, we conduct a preliminary analysis on two evaluation tasks: Relation Extraction (RE) and Positive Relation Classification (PRC).

\paragraph{Settings of Our Analysis}
Consider a text $T$ and a set of $n$ entities $\{e_1, ..., e_n\}$. $e_i$ and $e_j$ denote two different entities, and $R$ is a set of predefined relations, including \textit{no\_relation}. The objective of the RE task is to identify the relation type in $R$ for each entity pair $(e_i, e_j)$.  
Under the PRC setting, we do not use any entity pairs of \textit{no\_relation}, and the model is trained and evaluated with only entity pairs that have some pre-defined relation types.
Such a setting allows us to assess the difficulty of differentiating the predefined relation types.
We use the top-performing models for our preliminary experiments, and we report the scores on the development sets in these experiments. For the sentence-level task, we use the typed entity marker \citep{zhou2021improved} with RoBERTa-large~\citep{liu2019roberta} as our baseline. 
For the document-level task, we use KD-DocRE \citep{tan2022document} as the baseline. We use RoBERTa-large as the encoder for DocRED and XLM-R-base~\citep{conneau2020unsupervised} for HacRED.  

\paragraph{Datasets}
We compare the preliminary investigation results on two sentence-level RE datasets: TACRED \citep{zhang-etal-2017-position} and Re-TACRED~\cite{stoica2021re}, and two document-level RE datasets: DocRED \citep{yao2019docred}  and HacRED~\citep{cheng-etal-2021-hacred}. Table \ref{tab:data-stats} shows the statistics of the datasets.
Re-TACRED revised the TACRED dataset to reduce annotation errors.
Besides, we also create an incomplete Re-TACRED dataset, where 30\% of the positive labels are replaced with \textit{no\_relation} in the training, development, and test sets. 
This modification artificially creates false negative samples. This is to show the detrimental effect of the incompleteness problem on downstream tasks.
DocRED and HacRED are document-level relation extraction datasets in English and Chinese, respectively.
For sentence-level relation extraction datasets, the negative ratio is typically lower than that of the document-level datasets, mainly because the complexity of relation extraction is quadratic in the number of entities. Therefore, document-level RE is more prone to the false negative problem.

\begin{table}[t!]
\centering
\resizebox{1\columnwidth}{!}{
\begin{tabular}{lccc} 
\toprule
 \textbf{Dataset}         & P     & R   & F1    \\ 
\midrule
\multicolumn{2}{l}{\textbf{Sentence-level}} \\ 
TACRED    & 93.38 & 93.38 & 93.38  \\
Re-TACRED & 96.83 & 96.83  & 96.83  \\
Incomplete Re-TACRED & 96.53 & 96.53  & 96.53 \\ 
\midrule
\multicolumn{2}{l}{\textbf{Document-level}}  \\ 
DocRED    & 93.33 & 90.28 & 91.78  \\
HacRED    & 95.18 & 94.95 & 95.07  \\
\bottomrule
\end{tabular}
}
\caption{Preliminary experimental results of PRC.}
\label{tab:pos-rel}
\end{table}


\subsubsection{Preliminary Results of Positive Relation Classification}
\label{sec:prc}
By assuming that there is a relation between the entity pairs, the positive relation classification (PRC) task shows the difficulty of differentiating the relation types. 
Note that document-level PRC is a multi-label classification problem. Hence, precision and recall are not necessarily the same.   
From Table~\ref{tab:pos-rel}, we can see that all the models perform well. 
The performance of sentence-level RE and document-level RE are comparable, even though document-level RE has a significantly longer context and requires cross-sentence reasoning. 
This shows that the difficulty of positive relation classification is not severely affected by sentence boundary or context length. 
Another finding from Table~\ref{tab:pos-rel} is that the revised version of TACRED has marginally higher performance than the original version. This is expected as the revised version receives an extra round of human annotation. 
The performance on Incomplete Re-TACRED is only marginally worse than Re-TACRED, which shows that positive relation classification can achieve a comparatively high performance despite the reduction of training instances. 
Besides, it is worth noting that the performance on HacRED is higher than the performance on DocRED, even though HacRED claims that it is a semantically harder dataset. 
Although the baselines are not exactly the same, we can still infer that the difficulty level of classifying the positive relation types on sentence-level and document-level datasets is not significantly different.

\begin{table}[t!]
\centering
\resizebox{1\columnwidth}{!}{
\begin{tabular}{lccc} 
\toprule
\textbf{Dataset} & P     & R    & F1     \\ 
\midrule
\multicolumn{2}{l}{\textbf{Sentence-level}}  \\ 
TACRED    & 75.70 & 	73.40 & 74.50  \\
Re-TACRED & 90.60	& 91.30   & 90.90 \\
Incomplete Re-TACRED & 65.61 & 71.71  & 68.52 
  \\ 
\midrule
\multicolumn{2}{l}{\textbf{Document-level}} \\
DocRED    & 64.62 & 63.53 & 64.07  \\
HacRED    & 78.45 & 77.93 & 78.19  \\
\bottomrule
\end{tabular}
}
\caption{Preliminary experimental results of RE.}
\label{tab:rel-extraction}
\end{table}

\subsubsection{{Preliminary Results of Relation Extraction}}

Compared to the setting of positive relation classification, the standard relation extraction task includes all the negative \textit{no\_relation} samples during training.
We compare the performance of the previous best approaches on the sentence-level and document-level RE datasets in Table~\ref{tab:rel-extraction}.
We observe that the performance on the standard RE task is lower than that on the PRC task. For a specific dataset, the performance on PRC is the upper bound of RE performance, since the evaluation of PRC ignores \textit{no\_relation}.
However, the performance between relation extraction and positive relation classification should not have a large gap, as the positive samples should be semantically different from the negative samples for a well-annotated dataset.
For the sentence-level dataset, we can see that the performance on the revised version of TACRED (Re-TACRED) is significantly higher than that of the original TACRED dataset.
The performance difference between the positive relation classification and relation extraction is 18.88 (93.38 vs. 74.50) F1 score for TACRED, whereas the gap is only 5.93 (96.83 vs. 90.90) F1 score for Re-TACRED. For the incomplete Re-TACRED, this gap becomes 28.01 (96.53 vs. 68.52). Moreover, from the comparison of Re-TACRED and incomplete Re-TACRED, we can see that the performance of PRC only dropped  by 0.3 (96.83 vs. 96.53), whereas the performance of RE dropped by 22.38 (90.90 vs. 68.52). This shows that the incompleteness problem greatly decreases RE performance even if the positive instances are precisely labeled, whereas the performance of PRC is not significantly affected by the incompleteness problem. 
For the document-level datasets, we observe large gaps between  positive relation classification and relation extraction, 27.71 (91.78 vs. 64.07) F1 for DocRED and 16.88 (95.07 vs. 78.19) for HacRED, respectively.
Compared to the 5.93 F1 gap on the Re-TACRED dataset, the gaps of PRC and RE on the document-level datasets are significantly larger. These observations indicate that there may exist a substantial number of false negative examples throughout the training, development, and test sets of these datasets.  
Therefore, it is necessary to re-check the quality of the document-level relation extraction datasets.

\section{Revising DocRED with an Iterative Approach}
\label{sec:revision}
In this section, we describe our iterative human-in-the-loop approach to revise DocRED.
The goal is to add the previously unrecognized relation triples to the DocRED dataset. 
Our iterative approach consists of three steps in each iteration: 
(1) Training scorer models;
(2) Scoring all possible triples; and
(3) Human verification.
The following sections provide the details of each step.

\begin{table}[ht]
\centering
\begin{tabular}{llll} 
\toprule
Split 0~ & Split 1 & Split 2 & Split 3  \\ 
\midrule
1,000  & 1,000 & 1,000 & 1,053  \\
\bottomrule
\end{tabular}
\caption{Dataset statistics of each split. Split 0 is the original DocRED development set.}
\label{tab:data-split}
\end{table}  

\subsection{Our Iterative Approach}
\label{sec:iterative}
\paragraph{Step 1 - Training Scorer Models}
\label{sec:step1}
Even though the annotation of the previous DocRED is incomplete, we can still train neural models on such data.
We adopt three top-performing models on the current DocRED leaderboard: KD-DocRE~\cite{tan2022document}, DocuNET~\citep{zhang2021document}, and ATLOP~\citep{zhou2021document}. 
To obtain relation triple candidates for all 4,053 documents, we split the original DocRED into 4 different splits (Table~\ref{tab:data-split}), with the first split as the original DocRED development set. This is to ensure that the number of training samples is comparable to the original DocRED when any three of the splits are used for training our scorer models, and the remaining split is used for prediction. Following the training paradigm of \citet{tan2022document}, we first pre-train each model with the distantly supervised data in DocRED. Then, the pre-trained model is further trained on any three splits of DocRED,
and we then use the trained model to make predictions on the remaining split. Therefore, it requires four rounds of training and inference so that we can get the predictions for all the four splits.


\paragraph{Step 2 - Scoring the Triples}
\label{sec:cand-generation}
In this step, we aim to generate a large number of relation triple candidates, so that they could cover the missing annotations in the previous DocRED.
With the trained scorer models, we can predict the scores for all the enumerated relation triples.
To control the number of relation triple candidates for the next step, we define a threshold score to remove the less confident predictions. Specifically, we set the dynamic threshold to 0.9 of the models for the Adaptive Threshold class \citep{zhou2021document} for all trained models. 
The predicted relation triples from all the models are then merged together.
Due to the different characteristics of these models, we could generate a large and diverse pool of relation triple candidates for the next step. By default, the original positive triples from DocRED are treated as correct and are not  re-annotated.

\paragraph{Step 3 - Human Verification}
After the relation triple candidates are generated from the previous step, each triple candidate will be annotated by humans. The human annotators are asked to read the document and check whether the triples can be inferred from the document. 
Each triple will be annotated by two annotators, and a third annotator will resolve the conflicting annotations. 




\begin{table}[t]
\centering

\resizebox{\columnwidth}{!}{
\begin{tabular}{lccccc} 
\toprule
                      \multirow{2}{*}{\textbf{Dataset}}    & \multicolumn{3}{c}{\textbf{Re-DocRED}}        & \multicolumn{2}{c}{\textbf{DocRED}}  \\
                          \cmidrule(lr){2-4} \cmidrule(lr){5-6}
                          & \textbf{Train} & \textbf{Dev} & \textbf{Test} & \textbf{Train} & \textbf{Dev}        \\ 
\midrule
\textbf{\# Documents}      & 3,053          & 500          & 500           & 3,053          & 1,000               \\
\textbf{Avg. \# Entities}  & 19.4           & 19.4         & 19.6          & 19.5           & 19.6                \\
\textbf{Avg. \# Triples}   & 28.1           & 34.6           & 34.9          & 12.5           & 12.3                \\
\textbf{Avg. \# Sentences} & 7.9            & 8.2          & 7.9           & 7.9            & 8.1                 \\
\bottomrule
\end{tabular}}
\caption{Statistics of Re-DocRED and DocRED. }
\label{tab:data-stats-doc}
\end{table}

\begin{table*}
\centering
\resizebox{0.9\textwidth}{!}{
\begin{tabular}{lcccccccccc} 
\toprule
\multirow{3}{*}{\textbf{Model}} & \multicolumn{4}{c}{\textbf{DocRED}} & \multicolumn{4}{c}{\textbf{Re-DocRED}} & \multicolumn{2}{c}{\multirow{2}{*}{\textbf{Test Differences}}}   \\
\cmidrule(lr){2-5} \cmidrule(lr){6-9}
\multicolumn{1}{l}{} & \multicolumn{2}{c}{\textbf{Dev}}                              & \multicolumn{2}{c}{\textbf{Test}}                             & \multicolumn{2}{c}{\textbf{Dev}}                              & \multicolumn{2}{c}{\textbf{Test}}                             &   \\
\cmidrule(lr){2-3} \cmidrule(lr){4-5}\cmidrule(lr){6-7} \cmidrule(lr){8-9} \cmidrule(lr){10-11}
& \multicolumn{1}{c}{Ign\_F1} & \multicolumn{1}{c}{F1} & \multicolumn{1}{c}{Ign\_F1} & \multicolumn{1}{c}{F1} & \multicolumn{1}{c}{Ign\_F1} & \multicolumn{1}{c}{F1} & \multicolumn{1}{c}{Ign\_F1} & \multicolumn{1}{c}{F1} & \multicolumn{1}{c}{$\Delta$Ign\_F1} & \multicolumn{1}{c}{$\Delta$F1}  \\ 
\midrule
JEREX                & 57.07	                       & 58.97	                 & 57.14	                       & 59.01                  &  71.59	                 & 72.68                & 	71.45                  & 72.57                     & $+$14.31	                             & $+$13.56                        \\
ATLOP                & 61.21	                       & 63.07	                 & 61.44                      & 	63.20                    & 76.79                     & 	77.46                  & 	76.82	                       & 77.56                  & $+$15.38	                            & $+$14.36                         \\
DocuNET              & 61.62	                     & 63.53	                  & 61.80	                      & 63.64                  & 77.49                     & 	78.14	                   & 77.26                       & 	77.87               & $+$15.46		                             & $+$14.23                           \\
KD-DocRE                & 62.08	                      & 64.07	                   & 62.21	                       & 64.07                   & 77.85                            & 	78.51       & 77.60               & 	78.28                           & $+$15.39	                             & $+$14.21 
  \\ 
\midrule
\multicolumn{6}{l}{\textbf{+ Pre-trained with distantly supervised data}}  \\JEREX                & 60.39	                       & 62.24	                 & 60.29	                        & 62.15                  & 73.34	                      &74.77	                   & 73.48                       & 	74.79                  & $+$13.19	                             & $+$12.64                         \\
ATLOP                & 63.68	                       & 65.61	                & 63.63	                        & 65.51                  & 78.32	  		         & 79.26	                  & 78.52	                    & 79.46	                   & $+$14.89	                           & $+$13.95                          \\
DocuNET              & 63.22                     & 	65.25	                   & 63.23                      & 	65.26                   & 78.20	                       & 78.90	                 & 78.28	                       & 78.99	                  & $+$15.06		                            & $+$13.73  \\
KD-DocRE                & 65.11                       & 	67.04	                  & 65.21                     & 	67.09                    & 79.79	                         & 	80.56		               & 80.32               & 		81.04	                 & $+$15.11			                           & $+$13.95                         \\
\bottomrule
\end{tabular}
}
\caption{Experimental results using the original DocRED and our Re-DocRED. The reported numbers are average scores over 3 runs. The reported DocRED results use the same splits of development and test sets as Re-DocRED.}
\label{tab:revised-exp-no-ds}
\end{table*}

\subsection{Our Revised Re-DocRED}
\label{sec:re-docred-stats}
The above three steps form one round of our iterative approach. We conducted two rounds of annotation in total. For the first round, we annotated 4,053 documents that include all training and evaluation documents. On average, we recommended 11.9 triples for each document and 9.4 triples were accepted, with an acceptance rate of 79.0\%. The Fleiss Kappa~\citep{fleiss1971measuring} coefficient for round 1 annotation is 0.73, which is considered as substantial agreement. To further improve the recall on the evaluation dataset, we conducted a second round of annotation for the 1,000 evaluation documents. We used the annotated 3,053 training samples from round 1 for round 2 training. 
In this round, 14.1 triples were recommended and only 6.0 triples were accepted, with an acceptance rate of only 42.6\%. The Fleiss Kappa coefficient for round 2 annotation is 0.66.
After human annotation, we also add relation triples by manually defining logical rules. In this way, we are able to resolve the problem of logical inconsistency (described in Section \ref{sec:problem-docred}) in the DocRED dataset. These rules mainly consist of inverse relations and co-occurring relations. See Appendix \ref{app:logic-detail} for more details.
Overall, our training documents contain 28.1 triples on average, with 9.4 triples added from human annotation and 6.2 triples from logical rules. Our evaluation documents contain 34.7 triples on average, with 15.7 triples added from human annotation and 6.7 triples from logical rules. We divide the 1,000 evaluation documents into 500 development and 500 test documents. The average number of triples of the evaluation documents is higher than that of the training documents. This indicates that the evaluation data has more complete annotation compared to the training data. The detailed statistics of the Re-DocRED dataset are shown in Table~\ref{tab:data-stats-doc}. The average number of triples per document is significantly higher for Re-DocRED compared to the original DocRED. There are 12.3 triples per document in the DocRED dev set and 34.7 triples per document in the dev and test sets of Re-DocRED. This shows that approximately 64.6\% of all triples are missing in the original DocRED dataset.


 \section{Experiments}

\subsection{Comparison on Relation Extraction}
\label{sec:experiment-compare}
To compare the previous DocRED and our Re-DocRED, we evaluate 4 different approaches on the two datasets.
Apart from the three models that are used during our annotation process (Section~\ref{sec:step1}), we also compare the performance with an additional approach, JEREX \cite{eberts2021end}. This approach is not included in our data production process and we use it as an independent model to compare DocRED and Re-DocRED.
Table~\ref{tab:revised-exp-no-ds} shows the experimental results, and the reported metrics are micro-averaged F1 scores and Ign\_F1 scores. The latter refers to the F1 score that ignores the triples appearing in the training set.
According to the statistics in Table \ref{tab:data-stats}, even though our revised Re-DocRED dataset contains many more relation triples, we observe that all the baseline models demonstrate significant performance improvement on both development and test sets. 
Compared to DocRED, the performance of the baseline models on Re-DocRED increased by more than 12 F1 points. 
When these models are pre-trained with distantly supervised data\footnote{{There are 101K distantly supervised documents in the original DocRED dataset.}}, we observe consistent performance improvement on Re-DocRED. We also further analyze the error cases of the SOTA model in Appendix~\ref{sec:model-mistakes}.


\begin{table}[t]
\centering
\resizebox{1\columnwidth}{!}{
\begin{tabular}{llccc} 
\toprule
\textbf{Training}     & \textbf{Test}      & P     & R    & F1     \\ 
\midrule
DocRED    & DocRED    & 93.33 & 90.28 & 91.78  \\
Re-DocRED & Re-DocRED & 95.03 & 90.55 & 92.74  \\
\bottomrule
\end{tabular}}
\caption{Positive relation classification performance with the original DocRED and our revised Re-DocRED (using KD-DocRE).}
\label{tab:pos-rc-rev}
\end{table}

\begin{table*}[ht]
\centering
\resizebox{0.8\textwidth}{!}{\begin{threeparttable}
\begin{tabular}{lcccccccc} 
\toprule
\multirow{2}{*}{\textbf{Model}}& \multicolumn{2}{c}{\textbf{Dev}} &\multicolumn{6}{c}{\textbf{Test}} \\
\cmidrule(lr){2-3}\cmidrule(lr){4-9}
& \textbf{Ign\_F1} & \textbf{F1}    & \textbf{Ign\_F1} & \textbf{F1}    & \textbf{Freq. F1} & \textbf{LT F1} & \textbf{Intra F1} & \textbf{Inter F1}  \\ 

\midrule
JEREX   & 71.59                            & 	72.68	           & 71.45                 & 	72.57               & 77.09		              & 66.31              & 		76.10	           & 	69.88                   \\
ATLOP   & 76.79                             & 	77.46	           & 76.82                  & 	77.56               & 80.78	              & 72.29               & 	80.11	           & 74.92                  \\
DocuNET & 77.49                 & 	78.14	           & 77.26	                           & 77.87                  & 81.25                & 73.32                & 	79.89	                & 76.58                 \\
KD-DocRE   & 77.85	                     & 78.51	         & 77.60                             &	78.28	                 & 80.85	               & 74.31	              & 79.52	                & 77.18   \\
\midrule
\multicolumn{6}{l}{\textbf{+ Pre-trained with distantly supervised data}}  \\
JEREX   & 73.34		                             &74.77		            & 73.48	               & 	74.79        & 	78.67	 	                      & 68.62		              & 76.90		               & 72.65                 \\
ATLOP   & 78.32	                             &79.26	            & 78.52                & 	79.46          & 	82.23	                      &75.17	              & 80.39	               & 78.44                  \\
DocuNET & 78.20	                              & 78.90	                & 78.28               & 	78.99	                 & 82.08                & 	74.19                &	80.46	              & 77.72                 \\
KD-DocRE   & 79.79                      & 	80.56                      & 	80.32	                & 81.04	                & 83.17	               & 79.31	                 & 82.01	                & 79.87      \\
\bottomrule
\end{tabular}
\end{threeparttable}
}
\caption{Performance comparison under different metrics on the Re-DocRED dataset. }
\label{tab:benchmark}
\end{table*}

\subsection{Comparison on Positive Relation Classification}
Following the experimental setting in Section \ref{sec:prc},  
we compare the positive relation classification performance between the original DocRED and our Re-DocRED with KD-DocRE~\citep{tan2022document} in Table~\ref{tab:pos-rc-rev}. 
The performance on Re-DocRED is slightly better than that on the original DocRED. 
This indicates that our added triples are of comparable quality to the original DocRED data.

\subsection{More Analysis}
\label{sec:analysis}

\paragraph{Additional Evaluation Metrics} As mentioned in Section \ref{sec:preliminaries}, document-level relation extraction is a challenging task. Hence, it is necessary to have various performance evaluation metrics so as to conduct a comprehensive evaluation.
On top of the standard F1 and Ign\_F1 evaluation metrics, we use four additional metrics to assess the models. 
(1) \textbf{Freq. F1}~\citep{tan2022document}, which only considers the 10 most common relation types in the training set of Re-DocRED, where these frequent relation types account for around 60\% of the relation triples. 
(2) \textbf{LT F1}, which only considers the long-tail (the remaining 86) relation types. 
(3) \textbf{Intra F1}~\citep{nan-etal-2020-reasoning}, which evaluates on relation triples that appear in the same sentence. 
(4) \textbf{Inter F1}, which evaluates on cross-sentence relation triples. 

We show the comprehensive evaluation results in Table~\ref{tab:benchmark}. We observe that there exists a relatively large gap between the \textbf{Freq. F1} and \textbf{LT F1} metrics, and the difference is around 6 to 10 F1 points. Such behavior shows that the frequent relation types are easier to be recognized compared to the long-tail relation types.
Furthermore, we also find that the performance on triples that appear in the same sentence (\textbf{Intra F1}) is better than that on the cross-sentence relation triples (\textbf{Inter F1}), by around 2 to 6 F1 points. This is because it is harder to encode long-distance interactions.
Therefore, future research can work on matching the performance of the long-tail relation types to the frequent types and also improve the model's representation capability to capture inter-sentence interactions.

\begin{table}[t]
\centering
\resizebox{1\columnwidth}{!}{
\begin{tabular}{lcccc} 
\toprule
\textbf{Training}     & \textbf{Negative Sample}      & P     & R    & F1     \\ 
\midrule
DocRED    & All  &  92.08 & 32.07 & 47.57 \\
Re-DocRED & All & 89.76 & 69.40  & 78.28  \\
\midrule
DocRED    & Partial (10\%)    & 77.96 & 51.97  & 62.37\\
Re-DocRED & Partial (70\%)  & 83.58 & 75.06 & 79.09  \\
\bottomrule
\end{tabular}}
\caption{Experimental results when using different numbers of negative samples. }
\label{tab:negative-sampling}
\end{table}

\paragraph{Training with DocRED vs. Re-DocRED}
We compare the performance of the KD-DocRE model when it is trained on the training sets of DocRED and Re-DocRED, and the results are shown in the upper section of Table \ref{tab:negative-sampling}. 
Note that for reliability, the evaluations are conducted on the development and test sets of Re-DocRED.
Under the default setting (using all the positive and negative samples during training), the model trained with the Re-DocRED training data achieves a comparable precision score as DocRED (89.76 vs. 92.08). 
However, the recall score of the model trained with Re-DocRED is much better than DocRED (69.40 vs. 32.07).
This demonstrates the severe incompleteness of the previous DocRED's annotation, and models trained with such data fail to capture many triples.
After adding back the missed samples, Re-DocRED's training set is more complete and thus the trained model tends to extract more valid triples.
Figure~\ref{fig:example-doc-pred} shows an example document with predictions of models trained on the two training sets. 
We can see that the model trained with Re-DocRED indeed outputs more correct triples.
On the other hand, for Re-DocRED, the recall score is still lower than the corresponding precision score. One possible reason is that the evaluation sets of Re-DocRED received two rounds of annotation while the training set only received one round. Therefore, the annotation of the evaluation sets of Re-DocRED is more complete compared to the training set.

To further compare the annotation quality, we conduct experiments to  randomly sample a subset of the negative samples instead of using all of them. This training strategy is known as negative sampling, and it is shown to be effective in resolving the false negative problem \citep{li2021empirical} during training for named entity recognition.
The experimental results on negative sampling are given in the lower section of Table~\ref{tab:negative-sampling}. 
When random down-sampling on negative entity pairs is adopted, the experimental results are improved, due to significantly improved recall.
We also observe that the improvement is more significant on DocRED. 
This demonstrates that the incompleteness problem of our revised training data is significantly alleviated. 
The sampling rate is a hyper-parameter obtained by grid search, and the details of negative sampling are described in Appendix~\ref{app:neg-sample-detail}.


\begin{figure}[!t]
\centering
\small
\begin{tabular}{|p{0.92\columnwidth}|}
\hline
\vspace{-.1cm}
\textbf{Ross Patterson Alger} ( \textbf{August 20 , 1920} – \textbf{January 16 , 1992} ) was a politician in the \textbf{Canadian} province of \textbf{Alberta} , who served as mayor of \textbf{Calgary} from 1977 to 1980 . Born in \textbf{Prelate} , \textbf{Saskatchewan} , he moved to \textbf{Alberta} with his family in 1930s . He received a bachelor of commerce degree from the \textbf{University of Alberta} in \textbf{1942} . \colorbox{LightYellow}{\underline{He served with the}} \textbf{Royal Canadian Air Force} during \textbf{World War II} . After the war , \colorbox{LightYellow}{\underline{he received an MBA from the}} \textbf{University of Toronto}...  \\
\vspace{0.05cm}
\textbf{Subject Entity}: \textbf{Ross Patterson Alger}\\
\textcolor{orange}{\textbf{DocRED}}: (\textit{place of birth},  \textbf{Prelate}), (\textit{date of birth}, \textbf{August 20 , 1920}), (\textit{educated at}, \textbf{University of Alberta})...  \\
\vspace{0.05cm}
\textcolor{skyblue}{\textbf{Re-DocRED}}: (\textit{place of birth}, \textbf{Prelate}), (\textit{date of birth}, \textbf{August 20 , 1920}), (\textit{educated at}, \textbf{University of Alberta}), \color{darkgreen}{(\textit{educated at}, \textbf{University of Toronto})}, \color{darkgreen}{(\textit{military branch}, \textbf{Royal Canadian Air Force})}... 
\vspace{.1cm}\\
\hline
\end{tabular}
\caption{Sample predictions of the model when it is trained on DocRED and Re-DocRED training sets, and the subject entity of the selected triples is ``Ross Patterson Alger''. 
The \colorbox{LightYellow}{\underline{highlighted evidence}} verifies that the model trained on Re-DocRED has captured more correct inter-sentence triples (\textcolor{darkgreen}{colored in green}) \color{black}{than DocRED}.}
\label{fig:example-doc-pred}
\end{figure}


\section{Related Work}

\paragraph{Relation Extraction}
There is a series of relation extraction datasets built over the past decades, and they have significantly advanced research on RE. 
ACE 2005 \citep{walker2006ace} and SemEval 2010 Task 8~\citep{hendrickx2010semeval} created two sentence-level RE datasets by human annotation. However, these two datasets have a relatively small number of relation types and instances. 
The New York Times corpus \citep{sandhaus2008new} is another common relation extraction dataset used in the literature \citep{riedel-etal-2013-relation,nayak-ng-2019-effective}. It has been used for joint entity and relation extraction by \citep{nayak-ng-2020}. 
The large-scale TACRED~\citep{zhang-etal-2017-position} dataset is created based on the 2009--2014 TAC knowledge base population (KBP) challenges and crowd-sourced human annotations. FewRel~\citep{han2018fewrel} and FewRel 2.0~\citep{gao-etal-2019-fewrel} have been proposed to study the transferability and few-shot capability of RE models. However, early relation extraction datasets mainly focus on sentence-level RE, whereas many relations can only be expressed by multiple sentences.
The document-level relation extraction task has been proposed to build RE systems that are able to extract relations from multiple entities and sentences. \citet{yao2019docred} have created the DocRED dataset by distant supervision from Wikipedia articles and the Wikidata knowledge base. They sampled 5,053 documents for human annotation. The annotation strategy of DocRED is mainly based on machine recommendation and human filtering. With a similar approach, \citet{cheng-etal-2021-hacred} have created a Chinese document-level RE dataset that focuses on hard relation cases. 

\paragraph{Machine-Assisted Data Generation}
Since labeled data is expensive to obtain for complex NLP tasks, there are many research works on generating automatically labeled data. Distant supervision is first used by \citet{mintz2009distant} to generate relation extraction data without human efforts. 
Prior work on automatic data generation mainly relies on rule-based pattern matching \citep{lehmann2015dbpedia} and web crawling~\citep{buck2014n}. These types of rule-based methods are susceptible to noise propagation. With the rapid development of pre-trained language models~(PLMs; \citealp{devlin-etal-2019-bert}; \citealp{liu2019roberta} ; \citealp{brown2020language}), 
many recent works leverage PLMs for automatic data generation~(\citealp{10.1145/2433396.2433468}; \citealp{anaby2020not};  \citealp{zhou-etal-2022-melm}; \citealp{yang-etal-2020-generative}; \citealp{kumar-etal-2020-data}; \citealp{liu-etal-2021-mulda}). 
However, these methods typically depend on a certain set of supervised data. Another line of work utilizes manually designed prompts and instructions to generate data in an unsupervised manner~(\citealp{wang-etal-2019-tackling}; \citealp{schick-schutze-2021-generating}; \citealp{chia2022relationprompt}). Although these methods improve the performance on some downstream tasks, the quality of the machine-generated data still does not match human annotation. To mitigate noise from the machine-generated data, \citet{west2021symbolic}  generate a large amount of commonsense knowledge data and employ human annotators to filter the generated candidates.

\section{Conclusions}
In conclusion, this paper identifies the causes and effects of the overwhelming false negative problem in relation extraction. We show that the false negative problem is the cause of the performance bottleneck on many RE datasets.
We have also provided a high-quality revised version of document-level RE dataset, namely Re-DocRED. Moreover, we have provided comprehensive experimental analysis for Re-DocRED and identified the potential areas for further improvement. We have also conducted a thorough error analysis on state-of-the-art RE models.

\section{Acknowledgements}
We would like to thank all our annotators for their efforts, and the anonymous reviewers for their insightful comments and feedback.

\section{Limitations}
In this section, we discuss the limitations of our work. We have revised a popular document-level relation extraction dataset -- the DocRED dataset. We use an iterative human-in-the-loop approach for revising this dataset. However, after two rounds of annotation of the evaluation data, there might still be missing annotations, even though we have added almost twice the number of triples compared to the original DocRED dataset. That is, our re-annotation serves as a reasonable approximation to the ground truth, although it may still not be the gold standard for an ideal document-level RE scenario. We have provided a detailed analysis on annotation quality and comparison with contemporaneous work in Appendix~\ref{app:comparisonwithscratch}. The second limitation of our work is that human revision on the training data is only carried out in one round. The completeness level of the training data is lower compared to our evaluation data. Future research can focus on advanced algorithms for learning from imperfectly annotated data, which is actually very common in real applications. Nevertheless, such study still needs high-quality test data for reliable evaluation. 

\section{Ethical Considerations}
This paper focuses on revising the annotation on the false negative examples in the DocRED dataset, a publicly available and widely used benchmark for document-level relation extraction. All the documents and relations are provided in the original DocRED dataset. The annotators involved in this work were paid around 60 CNY per hour, which is more than three times the minimum wage in that area. The scope of this work is on revising the relation triples in these documents. However, there may be improper content within the Wikipedia articles themselves. The authors of this paper are not responsible for ethical issues arising from such improper Wikipedia content. 

\bibliographystyle{acl_natbib}
\bibliography{anthology}

\begin{thebibliography}{47}
\expandafter\ifx\csname natexlab\endcsname\relax\def\natexlab#1{#1}\fi

\bibitem[{Alt et~al.(2020)Alt, Gabryszak, and Hennig}]{alt2020tacred}
Christoph Alt, Aleksandra Gabryszak, and Leonhard Hennig. 2020.
\newblock \href {https://aclanthology.org/2020.acl-main.142/} {{TACRED}
  revisited: A thorough evaluation of the {TACRED} relation extraction task}.
\newblock In \emph{Proceedings of ACL}.

\bibitem[{Anaby-Tavor et~al.(2020)Anaby-Tavor, Carmeli, Goldbraich, Kantor,
  Kour, Shlomov, Tepper, and Zwerdling}]{anaby2020not}
Ateret Anaby-Tavor, Boaz Carmeli, Esther Goldbraich, Amir Kantor, George Kour,
  Segev Shlomov, Naama Tepper, and Naama Zwerdling. 2020.
\newblock \href {https://ojs.aaai.org/index.php/AAAI/article/view/6233} {Do not
  have enough data? deep learning to the rescue!}
\newblock In \emph{Proceedings of AAAI}.

\bibitem[{Bing et~al.(2015)Bing, Chaudhari, Wang, and
  Cohen}]{bing-etal-2015-improving}
Lidong Bing, Sneha Chaudhari, Richard Wang, and William Cohen. 2015.
\newblock \href {https://aclanthology.org/D15-1060} {Improving distant
  supervision for information extraction using label propagation through
  lists}.
\newblock In \emph{Proceedings of EMNLP}.

\bibitem[{Bing et~al.(2013)Bing, Lam, and Wong}]{10.1145/2433396.2433468}
Lidong Bing, Wai Lam, and Tak-Lam Wong. 2013.
\newblock \href {https://doi.org/10.1145/2433396.2433468} {Wikipedia entity
  expansion and attribute extraction from the web using semi-supervised
  learning}.
\newblock In \emph{Proceedings of WSDM}.

\bibitem[{Bollacker et~al.(2008)Bollacker, Evans, Paritosh, Sturge, and
  Taylor}]{bollacker2008freebase}
Kurt Bollacker, Colin Evans, Praveen Paritosh, Tim Sturge, and Jamie Taylor.
  2008.
\newblock \href {https://dl.acm.org/doi/10.1145/1376616.1376746} {Freebase: a
  collaboratively created graph database for structuring human knowledge}.
\newblock In \emph{Proceedings of ACM SIGMOD}.

\bibitem[{Brown et~al.(2020)Brown, Mann, Ryder, Subbiah, Kaplan, Dhariwal,
  Neelakantan, Shyam, Sastry, Askell et~al.}]{brown2020language}
Tom Brown, Benjamin Mann, Nick Ryder, Melanie Subbiah, Jared~D Kaplan, Prafulla
  Dhariwal, Arvind Neelakantan, Pranav Shyam, Girish Sastry, Amanda Askell,
  et~al. 2020.
\newblock \href
  {https://proceedings.neurips.cc/paper/2020/file/1457c0d6bfcb4967418bfb8ac142f64a-Paper.pdf}
  {Language models are few-shot learners}.
\newblock In \emph{Proceedings of NIPS}.

\bibitem[{Buck et~al.(2014)Buck, Heafield, and Van~Ooyen}]{buck2014n}
Christian Buck, Kenneth Heafield, and Bas Van~Ooyen. 2014.
\newblock \href {https://aclanthology.org/L14-1074/} {N-gram counts and
  language models from the common crawl}.
\newblock In \emph{Proceedings of LREC}.

\bibitem[{Chen et~al.(2021)Chen, Fu, Lee, and Ma}]{chen2021h}
Jhih-wei Chen, Tsu-Jui Fu, Chen-Kang Lee, and Wei-Yun Ma. 2021.
\newblock \href {https://aclanthology.org/2021.findings-acl.228.pdf} {{H-FND:}
  hierarchical false-negative denoising for distant supervision relation
  extraction}.
\newblock In \emph{Findings of ACL}.

\bibitem[{Cheng et~al.(2021)Cheng, Liu, Qu, Zhao, Liang, Wang, Huai, Yuan, and
  Xiao}]{cheng-etal-2021-hacred}
Qiao Cheng, Juntao Liu, Xiaoye Qu, Jin Zhao, Jiaqing Liang, Zhefeng Wang,
  Baoxing Huai, Nicholas~Jing Yuan, and Yanghua Xiao. 2021.
\newblock \href {https://aclanthology.org/2021.findings-acl.249} {{H}ac{RED}: A
  large-scale relation extraction dataset toward hard cases in practical
  applications}.
\newblock In \emph{Findings of ACL}.

\bibitem[{Chia et~al.(2022)Chia, Bing, Poria, and Si}]{chia2022relationprompt}
Yew~Ken Chia, Lidong Bing, Soujanya Poria, and Luo Si. 2022.
\newblock \href {https://aclanthology.org/2022.findings-acl.5.pdf}
  {{RelationPrompt:} leveraging prompts to generate synthetic data for
  zero-shot relation triplet extraction}.
\newblock \emph{Findings of ACL}.

\bibitem[{Conneau et~al.(2020)Conneau, Khandelwal, Goyal, Chaudhary, Wenzek,
  Guzm{\'a}n, Grave, Ott, Zettlemoyer, and Stoyanov}]{conneau2020unsupervised}
Alexis Conneau, Kartikay Khandelwal, Naman Goyal, Vishrav Chaudhary, Guillaume
  Wenzek, Francisco Guzm{\'a}n, Edouard Grave, Myle Ott, Luke Zettlemoyer, and
  Veselin Stoyanov. 2020.
\newblock \href {https://aclanthology.org/2020.acl-main.747.pdf} {Unsupervised
  cross-lingual representation learning at scale}.
\newblock In \emph{Proceedings of ACL}.

\bibitem[{Devlin et~al.(2019)Devlin, Chang, Lee, and
  Toutanova}]{devlin-etal-2019-bert}
Jacob Devlin, Ming-Wei Chang, Kenton Lee, and Kristina Toutanova. 2019.
\newblock \href {https://aclanthology.org/N19-1423} {{BERT}: Pre-training of
  deep bidirectional transformers for language understanding}.
\newblock In \emph{Proceedings of NAACL}.

\bibitem[{Eberts and Ulges(2021)}]{eberts2021end}
Markus Eberts and Adrian Ulges. 2021.
\newblock \href {https://aclanthology.org/2021.eacl-main.319/} {An end-to-end
  model for entity-level relation extraction using multi-instance learning}.
\newblock In \emph{Proceedings of EACL}.

\bibitem[{Fleiss(1971)}]{fleiss1971measuring}
Joseph~L Fleiss. 1971.
\newblock \href {https://psycnet.apa.org/record/1972-05083-001} {Measuring
  nominal scale agreement among many raters.}
\newblock \emph{Psychological Bulletin}.

\bibitem[{Gao et~al.(2021)Gao, Han, Bai, Qiu, Xie, Lin, Liu, Li, Sun, and
  Zhou}]{gao2021manual}
Tianyu Gao, Xu~Han, Yuzhuo Bai, Keyue Qiu, Zhiyu Xie, Yankai Lin, Zhiyuan Liu,
  Peng Li, Maosong Sun, and Jie Zhou. 2021.
\newblock \href {https://aclanthology.org/2021.findings-acl.112.pdf} {Manual
  evaluation matters: Reviewing test protocols of distantly supervised relation
  extraction}.
\newblock In \emph{Findings of ACL}.

\bibitem[{Gao et~al.(2019)Gao, Han, Zhu, Liu, Li, Sun, and
  Zhou}]{gao-etal-2019-fewrel}
Tianyu Gao, Xu~Han, Hao Zhu, Zhiyuan Liu, Peng Li, Maosong Sun, and Jie Zhou.
  2019.
\newblock \href {https://doi.org/10.18653/v1/D19-1649} {{F}ew{R}el 2.0: Towards
  more challenging few-shot relation classification}.
\newblock In \emph{Proceedings of EMNLP}.

\bibitem[{Han et~al.(2020)Han, Gao, Lin, Peng, Yang, Xiao, Liu, Li, Zhou, and
  Sun}]{han-etal-2020-data}
Xu~Han, Tianyu Gao, Yankai Lin, Hao Peng, Yaoliang Yang, Chaojun Xiao, Zhiyuan
  Liu, Peng Li, Jie Zhou, and Maosong Sun. 2020.
\newblock \href {https://aclanthology.org/2020.aacl-main.75} {More data, more
  relations, more context and more openness: A review and outlook for relation
  extraction}.
\newblock In \emph{Proceedings of AACL}.

\bibitem[{Han et~al.(2018)Han, Zhu, Yu, Wang, Yao, Liu, and
  Sun}]{han2018fewrel}
Xu~Han, Hao Zhu, Pengfei Yu, Ziyun Wang, Yuan Yao, Zhiyuan Liu, and Maosong
  Sun. 2018.
\newblock \href {https://www.aclweb.org/anthology/D18-1514} {Fewrel: A
  large-scale supervised few-shot relation classification dataset with
  state-of-the-art evaluation}.
\newblock In \emph{Proceedings of EMNLP}.

\bibitem[{Hao et~al.(2021)Hao, Yu, and Hu}]{hao-etal-2021-knowing}
Kailong Hao, Botao Yu, and Wei Hu. 2021.
\newblock \href {https://aclanthology.org/2021.emnlp-main.761} {Knowing false
  negatives: An adversarial training method for distantly supervised relation
  extraction}.
\newblock In \emph{Proceedings of EMNLP}.

\bibitem[{Hendrickx et~al.(2010)Hendrickx, Kim, Kozareva, Nakov, S{\'e}aghdha,
  Pad{\'o}, Pennacchiotti, Romano, and Szpakowicz}]{hendrickx2010semeval}
Iris Hendrickx, Su~Nam Kim, Zornitsa Kozareva, Preslav Nakov, Diarmuid~{\'O}
  S{\'e}aghdha, Sebastian Pad{\'o}, Marco Pennacchiotti, Lorenza Romano, and
  Stan Szpakowicz. 2010.
\newblock \href {https://aclanthology.org/S10-1006.pdf} {Semeval-2010 task 8:
  Multi-way classification of semantic relations between pairs of nominals}.
\newblock In \emph{Proceedings of the 5th International Workshop on Semantic
  Evaluation}.

\bibitem[{Huang et~al.(2022)Huang, Hao, Ye, Zhu, Feng, and
  Zhao}]{Huang2022DoesRP}
Quzhe Huang, Shibo Hao, Yuan Ye, Shengqi Zhu, Yansong Feng, and Dongyan Zhao.
  2022.
\newblock \href {https://aclanthology.org/2022.acl-long.432/} {Does
  recommend-revise produce reliable annotations? an analysis on missing
  instances in {DocRED}}.
\newblock In \emph{Proceedings of ACL}.

\bibitem[{Kumar et~al.(2020)Kumar, Choudhary, and Cho}]{kumar-etal-2020-data}
Varun Kumar, Ashutosh Choudhary, and Eunah Cho. 2020.
\newblock \href {https://aclanthology.org/2020.lifelongnlp-1.3} {Data
  augmentation using pre-trained transformer models}.
\newblock In \emph{Proceedings of the 2nd Workshop on Life-long Learning for
  Spoken Language Systems}.

\bibitem[{Lehmann et~al.(2015)Lehmann, Isele, Jakob, Jentzsch, Kontokostas,
  Mendes, Hellmann, Morsey, Van~Kleef, Auer et~al.}]{lehmann2015dbpedia}
Jens Lehmann, Robert Isele, Max Jakob, Anja Jentzsch, Dimitris Kontokostas,
  Pablo~N Mendes, Sebastian Hellmann, Mohamed Morsey, Patrick Van~Kleef,
  S{\"o}ren Auer, et~al. 2015.
\newblock \href {http://svn.aksw.org/papers/2013/SWJ_DBpedia/public.pdf}
  {Dbpedia -- a large-scale, multilingual knowledge base extracted from
  wikipedia}.
\newblock \emph{Semantic web}.

\bibitem[{Li et~al.(2021)Li, Liu, and Shi}]{li2021empirical}
Yangming Li, Lemao Liu, and Shuming Shi. 2021.
\newblock \href {https://openreview.net/forum?id=5jRVa89sZk} {Empirical
  analysis of unlabeled entity problem in named entity recognition}.
\newblock In \emph{Proceedings of ICLR}.

\bibitem[{Liu et~al.(2021)Liu, Ding, Bing, Joty, Si, and
  Miao}]{liu-etal-2021-mulda}
Linlin Liu, Bosheng Ding, Lidong Bing, Shafiq Joty, Luo Si, and Chunyan Miao.
  2021.
\newblock \href {https://aclanthology.org/2021.acl-long.453} {{M}ul{DA}: A
  multilingual data augmentation framework for low-resource cross-lingual
  {NER}}.
\newblock In \emph{Proceedings of ACL}.

\bibitem[{Liu et~al.(2019)Liu, Ott, Goyal, Du, Joshi, Chen, Levy, Lewis,
  Zettlemoyer, and Stoyanov}]{liu2019roberta}
Yinhan Liu, Myle Ott, Naman Goyal, Jingfei Du, Mandar Joshi, Danqi Chen, Omer
  Levy, Mike Lewis, Luke Zettlemoyer, and Veselin Stoyanov. 2019.
\newblock \href {https://arxiv.org/abs/1907.11692} {Roberta: A robustly
  optimized bert pretraining approach}.
\newblock \emph{arXiv preprint arXiv:1907.11692}.

\bibitem[{Mintz et~al.(2009)Mintz, Bills, Snow, and
  Jurafsky}]{mintz2009distant}
Mike Mintz, Steven Bills, Rion Snow, and Dan Jurafsky. 2009.
\newblock \href {https://www.aclweb.org/anthology/P09-1113.pdf} {Distant
  supervision for relation extraction without labeled data}.
\newblock In \emph{Proceedings of ACL}.

\bibitem[{Nan et~al.(2020)Nan, Guo, Sekulic, and Lu}]{nan-etal-2020-reasoning}
Guoshun Nan, Zhijiang Guo, Ivan Sekulic, and Wei Lu. 2020.
\newblock \href {https://aclanthology.org/2020.acl-main.141} {Reasoning with
  latent structure refinement for document-level relation extraction}.
\newblock In \emph{Proceedings of ACL}.

\bibitem[{Nayak and Ng(2019)}]{nayak-ng-2019-effective}
Tapas Nayak and Hwee~Tou Ng. 2019.
\newblock \href {https://aclanthology.org/K19-1056} {Effective attention
  modeling for neural relation extraction}.
\newblock In \emph{Proceedings of CoNLL}.

\bibitem[{Nayak and Ng(2020)}]{nayak-ng-2020}
Tapas Nayak and Hwee~Tou Ng. 2020.
\newblock \href {https://ojs.aaai.org//index.php/AAAI/article/view/6374}
  {Effective modeling of encoder-decoder architecture for joint entity and
  relation extraction}.
\newblock In \emph{Proceedings of AAAI}.

\bibitem[{Riedel et~al.(2013)Riedel, Yao, McCallum, and
  Marlin}]{riedel-etal-2013-relation}
Sebastian Riedel, Limin Yao, Andrew McCallum, and Benjamin~M. Marlin. 2013.
\newblock \href {https://aclanthology.org/N13-1008} {Relation extraction with
  matrix factorization and universal schemas}.
\newblock In \emph{Proceedings of NAACL}.

\bibitem[{Sandhaus(2008)}]{sandhaus2008new}
Evan Sandhaus. 2008.
\newblock \href {https://catalog.ldc.upenn.edu/LDC2008T19} {The {New York
  Times} annotated corpus}.
\newblock \emph{Linguistic Data Consortium}.

\bibitem[{Schick and Sch{\"u}tze(2021)}]{schick-schutze-2021-generating}
Timo Schick and Hinrich Sch{\"u}tze. 2021.
\newblock \href {https://aclanthology.org/2021.emnlp-main.555} {Generating
  datasets with pretrained language models}.
\newblock In \emph{Proceedings of EMNLP}.

\bibitem[{Stoica et~al.(2021)Stoica, Platanios, and P{\'o}czos}]{stoica2021re}
George Stoica, Emmanouil~Antonios Platanios, and Barnab{\'a}s P{\'o}czos. 2021.
\newblock \href {https://ojs.aaai.org/index.php/AAAI/article/view/17631/17438}
  {{Re-tacred:} addressing shortcomings of the tacred dataset}.
\newblock In \emph{Proceedings of AAAI}.

\bibitem[{Tan et~al.(2022)Tan, He, Bing, and Ng}]{tan2022document}
Qingyu Tan, Ruidan He, Lidong Bing, and Hwee~Tou Ng. 2022.
\newblock \href {https://aclanthology.org/2022.findings-acl.132.pdf}
  {Document-level relation extraction with adaptive focal loss and knowledge
  distillation}.
\newblock In \emph{Findings of ACL}.

\bibitem[{Vrande{\v{c}}i{\'c} and Kr{\"o}tzsch(2014)}]{vrandevcic2014wikidata}
Denny Vrande{\v{c}}i{\'c} and Markus Kr{\"o}tzsch. 2014.
\newblock \href {https://dl.acm.org/doi/10.1145/2629489} {Wikidata: a free
  collaborative knowledgebase}.
\newblock \emph{Communications of the ACM}.

\bibitem[{Walker et~al.(2006)Walker, Strassel, Medero, and
  Maeda}]{walker2006ace}
Christopher Walker, Stephanie Strassel, Julie Medero, and Kazuaki Maeda. 2006.
\newblock \href {https://catalog.ldc.upenn.edu/LDC2006T06} {{ACE} 2005
  multilingual training corpus}.
\newblock \emph{Linguistic Data Consortium}.

\bibitem[{Wang et~al.(2019)Wang, Lai, Li, Bing, and
  Lam}]{wang-etal-2019-tackling}
Zihao Wang, Kwunping Lai, Piji Li, Lidong Bing, and Wai Lam. 2019.
\newblock \href {https://aclanthology.org/D19-1024} {Tackling long-tailed
  relations and uncommon entities in knowledge graph completion}.
\newblock In \emph{Proceedings of EMNLP}.

\bibitem[{West et~al.(2021)West, Bhagavatula, Hessel, Hwang, Jiang, Bras, Lu,
  Welleck, and Choi}]{west2021symbolic}
Peter West, Chandra Bhagavatula, Jack Hessel, Jena~D Hwang, Liwei Jiang,
  Ronan~Le Bras, Ximing Lu, Sean Welleck, and Yejin Choi. 2021.
\newblock \href {https://arxiv.org/pdf/2110.07178.pdf} {Symbolic knowledge
  distillation: from general language models to commonsense models}.
\newblock \emph{arXiv preprint arXiv:2110.07178}.

\bibitem[{Xiao et~al.(2020)Xiao, Yao, Xie, Han, Liu, Sun, Lin, and
  Lin}]{xiao-etal-2020-denoising}
Chaojun Xiao, Yuan Yao, Ruobing Xie, Xu~Han, Zhiyuan Liu, Maosong Sun, Fen Lin,
  and Leyu Lin. 2020.
\newblock \href {https://aclanthology.org/2020.emnlp-main.300} {Denoising
  relation extraction from document-level distant supervision}.
\newblock In \emph{Proceedings EMNLP}.

\bibitem[{Yang et~al.(2020)Yang, Malaviya, Fernandez, Swayamdipta, Le~Bras,
  Wang, Bhagavatula, Choi, and Downey}]{yang-etal-2020-generative}
Yiben Yang, Chaitanya Malaviya, Jared Fernandez, Swabha Swayamdipta, Ronan
  Le~Bras, Ji-Ping Wang, Chandra Bhagavatula, Yejin Choi, and Doug Downey.
  2020.
\newblock \href {https://aclanthology.org/2020.findings-emnlp.90} {Generative
  data augmentation for commonsense reasoning}.
\newblock In \emph{Findings of EMNLP}.

\bibitem[{Yao et~al.(2019)Yao, Ye, Li, Han, Lin, Liu, Liu, Huang, Zhou, and
  Sun}]{yao2019docred}
Yuan Yao, Deming Ye, Peng Li, Xu~Han, Yankai Lin, Zhenghao Liu, Zhiyuan Liu,
  Lixin Huang, Jie Zhou, and Maosong Sun. 2019.
\newblock \href {https://aclanthology.org/P19-1074/} {{DocRED:} a large-scale
  document-level relation extraction dataset}.
\newblock In \emph{Proceedings of ACL}.

\bibitem[{Zhang et~al.(2021)Zhang, Chen, Xie, Deng, Tan, Chen, Huang, Si, and
  Chen}]{zhang2021document}
Ningyu Zhang, Xiang Chen, Xin Xie, Shumin Deng, Chuanqi Tan, Mosha Chen, Fei
  Huang, Luo Si, and Huajun Chen. 2021.
\newblock \href {https://www.ijcai.org/proceedings/2021/0551.pdf}
  {Document-level relation extraction as semantic segmentation}.
\newblock \emph{Proceedings of IJCAI}.

\bibitem[{Zhang et~al.(2017)Zhang, Zhong, Chen, Angeli, and
  Manning}]{zhang-etal-2017-position}
Yuhao Zhang, Victor Zhong, Danqi Chen, Gabor Angeli, and Christopher~D.
  Manning. 2017.
\newblock \href {https://aclanthology.org/D17-1004} {Position-aware attention
  and supervised data improve slot filling}.
\newblock In \emph{Proceedings of EMNLP}.

\bibitem[{Zhou et~al.(2022)Zhou, Li, He, Bing, Cambria, Si, and
  Miao}]{zhou-etal-2022-melm}
Ran Zhou, Xin Li, Ruidan He, Lidong Bing, Erik Cambria, Luo Si, and Chunyan
  Miao. 2022.
\newblock \href {https://doi.org/10.18653/v1/2022.acl-long.160} {{MELM}: Data
  augmentation with masked entity language modeling for low-resource {NER}}.
\newblock In \emph{Proceedings of ACL}.

\bibitem[{Zhou and Chen(2021)}]{zhou2021improved}
Wenxuan Zhou and Muhao Chen. 2021.
\newblock \href {https://arxiv.org/abs/2102.01373} {An improved baseline for
  sentence-level relation extraction}.
\newblock \emph{arXiv preprint arXiv:2102.01373}.

\bibitem[{Zhou et~al.(2021)Zhou, Huang, Ma, and Huang}]{zhou2021document}
Wenxuan Zhou, Kevin Huang, Tengyu Ma, and Jing Huang. 2021.
\newblock \href {https://ojs.aaai.org/index.php/AAAI/article/view/17717/17524}
  {Document-level relation extraction with adaptive thresholding and localized
  context pooling}.
\newblock In \emph{Proceedings of AAAI}.

\end{thebibliography}
\appendix

\clearpage
\newpage

\begin{table*}[t]
\centering
        \resizebox{\textwidth}{!}{
        \begin{tabular}{p{22cm}}
            \toprule
            \textit{\textbf{Example 1}}\quad\quad\textit{\textbf{Error Type 1: Misunderstanding of Definition}: creator vs. architect}\\
            \textbf{Error Cause}: Annotator's Misunderstanding of Wikidata Relation Definition
            
            \textbf{Error Analysis}: There are over 9,000 relation types in the Wikidata knowledge base, whereas DocRED only contains 84 of them. The relation between buildings and its designer is architect (P84). But this relation is not in the DocRED's label space. In this case, the architect relation shall be excluded from DocRED. However, annotators from \citet{Huang2022DoesRP} uses creator to describe this kind of relation, which is not precise.
            \quad\quad

             \textbf{Document}: 
            Sir \colorB{\textbf{David Alan Chipperfield}} ( born 18 December 1953 ) is an English architect . He established David Chipperfield Architects in 1985 . His major works include the \colorB{\textbf{River and Rowing Museum}} in Henley - on - Thames , Oxfordshire ( 1989‚ 1998 ) ; the \colorB{\textbf{Museum of Modern Literature in Marbach}} , Germany ; the \colorB{\textbf{Des Moines Public Library}} , Iowa ( 2002‚ 2006 ) ; the \colorB{\textbf{Neues Museum}} , Berlin ( 1997 ‚ 2009 ) ; The \colorB{\textbf{Hepworth Wakefield gallery}} in Wakefield , UK ( 2003‚ 2011 ) , the \colorB{\textbf{Saint Louis Art Museum}} , Missouri ( 2005‚ 2013 ) ; and the \colorB{\textbf{Museo Jumex}} in Mexico City ( 2009‚ 2013 ) . Rowan Moore , the architecture critic of the Guardian of London , described his work as serious , solid , not flamboyant or radical , but comfortable with the history and culture of its setting . " He deals in dignity , in gravitas , in memory and in art . " David Chipperfield Architects is a global architectural practice with offices in London , Berlin , Milan , and Shanghai .
             \\ 
             \\
             \textbf{Wrong Triples}: (\colorB{\textbf{Museo Jumex}}, \colorG{\textit{creator}}, \colorB{\textbf{David Alan Chipperfield}}), (\colorB{River and Rowing Museum}, \colorG{\textit{creator}}, \colorB{\textbf{David Alan Chipperfield}}), (\colorB{\textbf{Museum of Modern Literature}}, \colorG{\textit{creator}}, \colorB{\textbf{David Alan Chipperfield}}), (\colorB{\textbf{Saint Louis Art Museum}},\colorG{\textit{creator}}, \colorB{\textbf{David Alan Chipperfield}}), (\colorB{\textbf{Hepworth Wakefield gallery}}, \colorG{\textit{creator}}, \colorB{\textbf{David Alan Chipperfield}}), (\colorB{\textbf{Neues Museum}}, \colorG{\textit{creator}}, \colorB{\textbf{David Alan Chipperfield}})

             \\

            \textit{\textbf{Example 2}}\quad\quad\textit{\textbf{Error Type 1: Misunderstanding of Definition}: series vs. part of}\\
            \textbf{Error Cause}: Annotator's Misunderstanding of Wikidata Relation Definition
            
             \textbf{Document}: \colorB{\textbf{Chapman Square}} is the debut studio album released by four piece British band Lawson . The album was released on 19 October 2012 via Polydor Records . The album includes their three top ten singles " When She Was Mine " , " \colorB{\textbf{Taking Over Me}} " and " \colorB{\textbf{Standing in the Dark}} " . The album was mainly produced by John Shanks with Duck Blackwell , Paddy Dalton , Ki Fitzgerald , Carl Falk , and Rami Yacoub . The album was re - released in the autumn of 2013 as \colorB{\textbf{Chapman Square Chapter II}} , with the lead single from the re - release being " Brokenhearted " , which features American rapper B.o . B. As of July 2016 , the album has sold 169,812 copies .
             \\ 
             \\
               \textbf{Wrong Triples}: (\colorB{\textbf{Taking Over Me}}, \colorG{\textit{series}}, \colorB{\textbf{Chapman Square Chapter II}}), (\colorB{\textbf{Brokenhearted}}, \colorG{\textit{series}}, \colorB{\textbf{Chapman Square Chapter II}}), (\colorB{\textbf{When She Was Mine}},~\colorG{\textit{series}}, \colorB{\textbf{Chapman Square}}), (\colorB{\textbf{Standing in the Dark}}, \colorG{\textit{series}}, \colorB{\textbf{Chapman Square}}) 
             \\
             
            \midrule
            \\
              \textit{\textbf{Example 3}}\quad\quad\textit{\textbf{Error Type 2: Commonsense Bias}}\\

             \textbf{Document}: 
            Alecu Russo ( born in March 17 , 1819 , near Chişinău , died on February 5 , 1859 , in \colorB{\textbf{Iaşi}} ) , was a \colorB{\textbf{Moldavian}} Romanian writer , literary critic and publicist . Russo is credited with having discovered one of the most elaborate forms of the Romanian national folk ballad Mioriţa . He was also a contributor to the Iaşi periodical Zimbrul , in which he published one of his best - known works , Studie Moldovană ( " Moldovan Studies " ) , in 1851 - 1852 . He also wrote Iaşii şi locuitorii lui în 1840 " Iaşi and its inhabitants in 1840 " – a glimpse into Moldavian society during the Organic Statute administration , and two travel accounts ( better described as folklore studies ) , Piatra Teiului and Stânca Corbului . Russo is also notable for his Amintiri ( " Recollections " ) , a memoir . 

            \\
            \textbf{Wrong Triples}: (\colorB{\textbf{Iaşi}}, \colorG{\textit{located in administrative territorial entity}}, \colorB{\textbf{Moldavian}}), (\colorB{\textbf{Moldavian}}, \colorG{\textit{contains administrative territorial entity}}, \colorB{\textbf{Iaşi}})
            \\
             \\
             \textit{\textbf{Example 4}}\quad\quad\textit{\textbf{Error Type 3: Slippery slope}}\\
            \textbf{Error Cause}: Improper reasoning based on punctuation for judgement of date of birth/death.
            
             \textbf{Document}: 
            South Wigston High School was founded in 1938 and is a school serving the local community of South Wigston .Today the school is an 11 – 16 yrs Academy. The main feeder primary schools are Glen Hills , Fairfield and Parkland . The school also attracts students from many areas of the city of Leicester and the county of Leicestershire . The school is oversubscribed and is growing year on year . South Wigston is known for its wide range of extra - curricular opportunities and for being a school that is inclusive and at the heart of the community . The school has extensive grounds and a purpose build sports centre opened by Gary Lineker . \colorbox{LightYellow}{Notable alumni} include \colorB{\textbf{Sue Townsend}} (  \colorB{\textbf{1946}} -  \colorB{\textbf{1950}} ) , author ; \colorB{\textbf{Louis Deacon}} ( \colorB{\textbf{1991}} - \colorB{\textbf{1995}} ) , Rugby Player for Leicester Tigers and England ; and Brett Deacon ( \colorB{\textbf{1992}} - \colorB{\textbf{1996}}) , Rugby Player for Leicester Tigers and England . In 2016 BBC2 produced a documentary entitled , The Secret Life of Sue Townsend Aged 68. Much of the documentary was filmed at the school and current students participated .
            \\
            \\
            \textbf{Wrong Triples}: (\colorB{\textbf{Louis Deacon}},  \colorG{\textit{date of birth}}, \colorB{\textbf{1991}}), (\colorB{\textbf{Louis Deacon}}, \colorG{\textit{date of death}}, \colorB{1995}), (\colorB{\textbf{Brett Deacon}}, \colorG{\textit{date of death}}, \colorB{\textbf{1996}}), (\colorB{\textbf{Brett Deacon}}, \colorG{\textit{date of birth}}, \colorB{\textbf{1992}}), (\colorB{\textbf{Sue Townsend}}, \colorG{\textit{date of birth}}, \colorB{1946}), (\colorB{\textbf{Sue Townsend}}, \colorG{\textit{date of death}}, \colorB{1950})
            \\
            \bottomrule
            
        \end{tabular}}
    \caption{Examples for the common error types by \citet{Huang2022DoesRP}. We use blue to color the \colorB{\textbf{entities}} and green to color the \colorG{\textit{relations}}. }
    \label{tab:error-case-scratch}
    \end{table*}   
\clearpage
\newpage

\section{Our Machine-Guided Annotation vs. Annotation from Scratch}
\label{app:comparisonwithscratch}
In this section, we compare our Re-DocRED dataset with a concurrent work \citep{Huang2022DoesRP} on revising the DocRED dataset. Our work uses machine-guided annotation methods, whereas their work asks the annotators to annotate from scratch (denoted as ``From Scratch''). As mentioned in \citet{Huang2022DoesRP}, annotating from scratch is an extremely challenging task. This is mainly due to the quadratic complexity of the document-level RE task. Suppose there are $N$ entities in one document and the label space of interest contains $R$ relations. The search space for human annotation is $N*(N-1)*R$. In particular, for an average case of DocRED ($N$=20, $R$=96), an annotator will need to make 36,480 classification decisions for one document. In contrast, for our machine-guided annotation, annotators will only need to make decisions on the recommended candidates, averaging only 25.5 decisions per document. This is primarily due to the pattern recognition capability of deep neural models, which significantly reduces the search space for human annotators. The size of our re-annotated dataset is much larger and our dataset contains all relations in the DocRED's label space, while the Scratch dataset only contains 91 out of the 96 relations in DocRED. Besides, we have also conducted significantly more experimental analysis.

It is also mentioned in \citet{Huang2022DoesRP} that re-annotation from scratch by two human experts may still not be the ground truth due to natural error rate. Therefore, we examined and analyzed all the annotated 96 documents of ``From Scratch''.
We found that there are several types of systematic errors in \citet{Huang2022DoesRP} and we show the error types in Table~\ref{tab:error-types-scratch}. 
Firstly, annotation from scratch is susceptible to annotators' misunderstanding of relation definition (25.5\%). For the first example in Table~\ref{tab:error-case-scratch}, the relation between an architect and his designed building is \textit{architect}\footnote{\url{https://www.wikidata.org/wiki/Property:P84}}, whereas annotators in \citet{Huang2022DoesRP} deem such relation as \textit{creator}\footnote{\url{https://www.wikidata.org/wiki/Property:P170}}. 
This is imprecise as the \textit{architect} relation was not in the label space of DocRED. Therefore, it is not possible to find this relation between the architect and its design. 
Secondly, annotating from scratch is susceptible to human commonsense bias (19.0\%). This is primarily due to human's memorization of popular entities, such as countries and geographical locations (example 3 in Table~\ref{tab:error-case-scratch}). 
The third major error type is due to slippery slope reasoning, as shown by example 4 in Table~\ref{tab:error-case-scratch}. 
The numbers within brackets were falsely identified as the \textit{date of birth} and \textit{date of death}, whereas the passage is about a renowned high school. 
It can be inferred that the numbers behind the alumni names are indicating the time periods that they were in this school. 
This error arises because most \textit{date of birth} and \textit{date of death} are described by brackets and numbers. 
However, such a pattern does not necessarily mean all numbers in brackets are indicating such relations.

\begin{table}[t]
\centering
\begin{tabular}{lc} 
\toprule
Error Types                    & Percentage  \\ 
\midrule
Misunderstanding of definition & 25.5\%      \\
Commonsense bias               & 19.0\%      \\
Slippery slope reasoning       & 22.1\%      \\
Others                         & 33.4\%      \\
\bottomrule
\end{tabular}
\caption{Common error types of \citet{Huang2022DoesRP}.}
\label{tab:error-types-scratch}
\end{table}

\begin{table}[t]
\centering
\resizebox{\columnwidth}{!}{
\begin{tabular}{lrrr} 
\toprule
     & \multicolumn{1}{l}{Added Triples} & \multicolumn{1}{l}{Errors} & \multicolumn{1}{l}{Precision}  \\ 
\midrule
From Scratch  & 2,057                               & 293                         & 85.8                          \\
Re-DocRED     & 2,067                               & 76                         & 96.3                           \\
\bottomrule
\end{tabular}}
\caption{Error rates of Re-DocRED and the From Scratch dataset~\citep{Huang2022DoesRP} based on examining all 96 documents. We observe that Re-DocRED has higher precision.}
\label{tab:compare-prec}
\end{table}

\begin{table}[t]
\centering
\resizebox{\columnwidth}{!}{
\begin{tabular}{lcc} 
\toprule
                        & Unit Price & Unit Time  \\ 
\midrule
Annotating from Scratch & 48 CNY     & 40 mins    \\
One Round of Revision     & 7.8 CNY    & 10 mins    \\
Two Rounds of Revision    & 15.8 CNY   & 15 mins    \\
\bottomrule
\end{tabular}}
\caption{Cost and unit time required for different annotation strategies.}
\label{tab:cost-analysis}
\end{table}

It is worth noting that the annotators in \citet{Huang2022DoesRP} are already experts in English and the annotators went through discussion after annotation. 
However, there are still a considerable number of errors from their dataset. 
We believe that this is due to the complex nature of this annotation task. 
In addition, we also conducted human evaluation of our Re-DocRED dataset and compared the precision of the two datasets in Table~\ref{tab:compare-prec}. We can see that our Re-DocRED dataset has significantly higher precision for the added triples. 
Moreover, we compare the unit price and unit time for different annotation strategies in Table \ref{tab:cost-analysis}. We can see that annotating from scratch costs three times more than our machine-guided annotation.

Hence, by comparing the two approaches for annotating document-level relation extraction datasets, we conclude that:

1. Even though ``From Scratch'' annotation is conducted  by human experts, there are still missing triples in the annotated 96 documents. That is, annotating from scratch does not completely eliminate the incompleteness problem when the number of entities $N$ and relation types $R$ are large. 

2. Annotation from scratch is not as precise as recommend-revise. As Table~\ref{tab:error-case-scratch} shows, human annotation of \citet{Huang2022DoesRP} contains several types of systematic errors.

3. Annotating from scratch is hard to scale. According to \citet{Huang2022DoesRP} and feedback from our annotators, it takes more than 30 minutes by experts to annotate one document. Then the two experts will still spend extra time discussing and resolving the conflicts.

4. The recommend-revise scheme is able to mitigate the false negative problem and is easier to scale up.

\begin{table*}[t]
\centering
        \resizebox{\textwidth}{!}{
        \begin{tabular}{p{22cm}}
            \toprule
            \textit{\textbf{Example 1}}\quad\quad\textit{\textbf{Error Type 1}: Extraneous Prediction (MR)}\\
            \textbf{Error Cause}: Popular pattern bias
            \quad\quad

             \textbf{Document}: 
            " Lookin Ass " is a song by American rapper and singer \colorB{\textbf{Nicki Minaj}} . It was produced by Detail and Choppa Boi . It was recorded by \colorB{\textbf{Minaj}} for the \colorB{\textbf{Young Money Entertainment}} compilation album ( 2014 ) . The music video for the track was released on February 14 , 2014. \textbf{...} On March 11 , 2014 , " Lookin Ass " was serviced to urban contemporary radio in the United States as Young Money : Rise of an Empires third official single . It was sent to US rhythmic radio stations on March 18 , 2014 , two weeks after its predecessor , " \colorB{\textbf{Trophies}} " .
             \\ 
             \\
             \textbf{Extraneous Triples}: (\colorB{\textbf{Young Money Entertainment}}, \colorG{\textit{performer}}, \colorB{Nicki Minaj}), (\colorB{\textbf{Trophies}}, \colorG{\textit{performer}}, \colorB{\textbf{Nicki Minaj}})

             \\

             \\

            \textit{\textbf{Example 2}}\quad\quad\textit{\textbf{Error Type 1}: Extraneous Prediction (MR)}\\
            \textbf{Error Cause}: Popular pattern bias
            \\
             \textbf{Document}: The Portland Golf Club is a private golf club in the northwest United States , in suburban \colorB{Portland} , Oregon . It is located in the unincorporated Raleigh Hills area of eastern \colorB{Washington County} , southwest of downtown \colorB{Portland} and east of Beaverton . PGC was established in the winter of 1914 , when a group of nine businessmen assembled to form a new club after leaving their respective clubs . The present site was chosen due to its relation to the Spokane , Portland and Seattle Railway 's interurban railroad line with frequent passenger service to the site because automobiles and roads were few . \textbf{...}
             \\ 
             \\
               \textbf{Extraneous Triples}: (\colorB{\textbf{Portland}},~\colorG{\textit{located in the administrative territorial entity}},~\colorB{\textbf{Washington County}}); (\colorB{\textbf{Washington County}}, \colorG{\textit{contains administrative territorial entity}}, \colorB{\textbf{Portland}})

             \\
                         \midrule
             \\
             \textit{\textbf{Example 3}}\quad\quad\textit{\textbf{Error Type 2}: Missing Triples (MS)}\\
            \textbf{Error Cause}: Failed in coreferential reasoning
            
             \textbf{Document}: 
            \colorB{Kurt Tucholsky} (; 9 January 1890 – 21 December 1935 ) was a German - Jewish journalist , satirist , and writer . He also wrote under the pseudonyms Kaspar Hauser ( after the historical figure ) , Peter Panter , Theobald Tiger and Ignaz Wrobel . Born in Berlin - Moabit , he moved to Paris in 1924 and then to Sweden in 1929 . \colorB{Tucholsky} was one of the most important journalists of the Weimar Republic . As a politically engaged journalist and temporary co - editor of the weekly magazine \colorB{Die Weltbühne} \colorB{he} proved himself to be a social critic in the tradition of Heinrich Heine . 
            \\
            \\
            \textbf{Missing Triples}: (\colorB{\textbf{Kurt Tucholsky}},  \colorG{\textit{employer}}, \colorB{\textbf{Die Weltbühne}})
            \\
            \\
              \textit{\textbf{Example 4}}\quad\quad\textit{\textbf{Error Type 2}: Missing Triples (MS)}\\

            \textbf{Error Cause}: Fail to find long-tail relations

             \textbf{Document}: 
            CBBC ( short for Children 's BBC ) is a British children 's television strand owned by the BBC and aimed for children aged from 6 to 12 . BBC programming aimed at under six year old children is broadcast on the CBeebies channel . CBBC broadcasts from 7   am to 9   pm on the digital CBBC Channel , available on most UK digital platforms . The CBBC brand was used for the broadcast of children 's programmes on BBC One on weekday afternoons and on BBC Two mornings until these strands were phased out in 2012 and 2013 respectively , as part of the BBC 's " Delivering Quality First " cost - cutting initiative . CBBC programmes were also broadcast in high definition alongside other BBC content on \colorB{\textbf{BBC HD}} , generally at afternoons on weekends , unless the channel was covering other events . This ended when \colorB{\textbf{BBC HD}} closed on \colorB{\textbf{26 March 2013}} , but CBBC HD launched on 10 December 2013 . 

            \\
            \textbf{Missing Triples}: (\colorB{\textbf{BBC HD}}, \colorG{\textit{dissolved, abolished or demolished}}, \colorB{\textbf{26 March 2013
            }})
            \\
            \bottomrule
            
        \end{tabular}
        }
    
    \caption{Examples of the two most common error types. We use blue and green to color the \colorB{\textbf{entities}} and \colorG{\textit{relations}}, respectively. }
    \label{tab:error-case}
    \end{table*}

\section{Coreferential Error Annotation}
\label{app:coreference}
\paragraph{Coreferential Errors} Besides the major problem of incomplete annotation, coreferential errors are also detrimental to the evaluation of DocRED. 
Note that an entity in the DocRED dataset can have multiple mention appearances in a document. 
If some mentions that are referring to the same entity are not included in the entity cluster, a redundant entity cluster will be formed. 
This kind of coreferential errors will affect the relation predictions involving the redundant entity. Since the complexity of the DocRE task is quadratic in the number of entities, it is important to make sure that the coreferential annotations are correct. Errors in coreferential annotation can be propagated and amplified during relation extraction. 

\paragraph{Coreferential Annotation} There are a considerable number of entities of DocRED that have the same surface names but refer to different entities. In such cases, we examined all entities that contain the same surface name and entity types in the DocRED dataset. For these entities, annotators will need to decide whether: (1)  the two entities are coreferential to each other, (2) the overlapping mentions are wrongly grouped to a certain entity cluster, and (3) the two mentions are indeed referring to two separate entities. As a result, we merged 102 coreferential entity pairs in the evaluation documents and 122 pairs in the training documents. Therefore, the average number of entities per document of Re-DocRED is slightly lower than the original DocRED.

\section{Details on Negative Sampling}
\label{app:neg-sample-detail}

\begin{figure}[ht]
    \centering
    \resizebox{0.7\columnwidth}{!}{
    \includegraphics{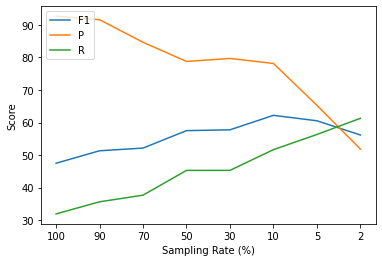}}
    \caption{Performance with respect to sampling rate. The model is trained on the training set of DocRED and is evaluated on the development set of Re-DocRED.}
    
    \label{fig:neg-samp-docred}
\end{figure}

\begin{figure}[ht]
    \centering
    \resizebox{0.7\columnwidth}{!}{
    \includegraphics{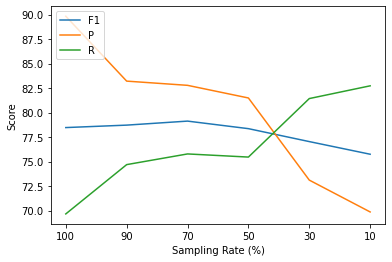}}
    \caption{Performance with respect to sampling rate. The model is trained and evaluated on Re-DocRED.}
    \label{fig:neg-samp-redocred}
\end{figure}

This section describes the details of the experiments with negative sampling. 
Given a text $T$ and a set of $n$ entities $\{e_1, ..., e_n\}$, the document-level RE task requires making predictions on all possible entity pairs $(e_i, e_j)$ for $i,j \in {1, ..., n}, i \neq j$. 
As we can see in Table~\ref{tab:data-stats}, over 90\% of the entity pairs are negative instances. Instead of using all negative instances, we sample a fraction of negative instances for training. 
Figure~\ref{fig:neg-samp-docred} shows the performance  with respect to the sampling rate when the model is trained on the training set of DocRED.
Figure~\ref{fig:neg-samp-redocred} shows the performance when the training set of Re-DocRED is used.
Note that evaluation is conducted on the development set of Re-DocRED.
From both figures, we observe that precision is positively correlated with the negative sampling percentage and recall is negatively correlated with this percentage. 
Besides, when training on highly incomplete data (DocRED), sampling 10\% of the negative instances improves the F1 score by 14.72 (62.21 vs 47.49) compared to using 100\% of the negative instances. 
However, the performance improvement is not that significant when training on Re-DocRED (79.13 vs 78.47), and the best sampling rate in this scenario is 70\%.

\section{Common Model Errors}
\label{sec:model-mistakes}
\begin{table}[t]
\newcommand{\multirot}[1]{\multirow{4}{*}[1.0ex]{\rotcell{\rlap{#1}}}}
\centering
\def\arraystretch{1.0}%
\resizebox{\columnwidth}{!}{
\begin{tabular}{l|l|l|c|}
\multicolumn{1}{l}{}          & \multicolumn{1}{l}{} & \multicolumn{2}{c}{\textbf{Ground Truth}}       \\ 
\cline{2-4}
\multirow{4}{*}{\rotcell{\clap{\textbf{Predictions}}}}&                      & $r \in$ \textbf{R} & \textbf{NA}                          \\ 
\cline{2-4}
                                &  \multirow{2}{*}{$r \in $ \textbf{R} }        & \textbf{C}: 13,610 (71.59\%)   & \multirow{2}{*}{\textbf{MR}: 1,884 (9.91\%)}  \\ 
\cline{3-3}
                                &            & \textbf{W}: 580 (3.05\%)     &                             \\ 
\cline{2-4}
                                & \textbf{NA}                   & \textbf{MS}: 2,935 (15.44\%)   &                     \\
\cline{2-4}
\end{tabular}}
\caption{Statistics of our error distribution on the development set of Re-DocRED based on KD-DocRE. The final evaluation score is evaluated on $r \in $ \textbf{R} triples, hence the correct predictions of \textbf{NA} are ignored when calculating the final scores.}
\label{tab:error-types}
\end{table}

In this section, we show examples of the mistakes predicted by the best model (KD-DocRE model with distant supervision pretraining) in our experiments.
Similar to \citet{tan2022document}, we split the union of ground-truth triples and predicted triples into four categories: (1) \textbf{Correct (C)}, where predicted triples are in the ground truth. (2) \textbf{Wrong (W)}, where the output relation type is wrong but the predicted head and tail entities are in the ground truth. (3) \textbf{Missed (MS)}, where the model predicts no relation for a pair of head and tail entities but there is some relation in the ground truth. (4) \textbf{More (MR)}, where the model predicts an extraneous relation for a pair of head and tail entities that is not in the ground truth. The performance breakdown is shown in Table~\ref{tab:error-types}. We observe that the majority of the errors were in the \textbf{MS} and \textbf{MR} categories. 
We found that predictions on popular relations tend to fall under the \textbf{MR} category. We further show this popularity bias pattern with examples in 
 Table~\ref{tab:error-case}.
We can see that popular relation patterns tend to be under the \textbf{MR} category. 
From the first example, ``Nicki Minaj'' is a popular artiste. But the song ``Trophies'' was not performed by ``Nicki Minaj''. In contrast, the model wrongly predicts that this song is also performed by her. On the other hand, in example 4, the cue that ``BBC HD'' dissolved on ``Mar 26, 2013'' is obvious, whereas the model failed to find this relation.

\section{Details of Logical Rules}
\label{app:logic-detail}

In this section, we show the logical rules that we used. 
After examining the DocRED dataset~\citep{yao2019docred}, we found that there are two types of logical inconsistencies. The first type is the incompleteness of inverse relations, and the second is the inconsistency in co-occurring relations. 
The inverse relations are logical relations that can be implied by reversing the direction of relation triples. 
For example, if entity 1 is the \textit{participant of} an event (entity 2), this event should have a \textit{participant} relation with entity 1. 
We show all the inverse relation pairs that we used in Table~\ref{tab:inverse-rel}. 
Besides inverse relations, we also added triples by co-occurring rules. 
This is mainly because we found that these relations are logically correlated and their co-occurrence is inconsistent in the original DocRED dataset. For example, when describing the relation between entities and wars, there are two involved relations: relation \textit{conflict}\footnote{\url{https://www.wikidata.org/wiki/Property:P607}} and \textit{participant of}\footnote{\url{https://www.wikidata.org/wiki/Property:P1344}}. For such cases, the two relations are considered to be present when \textit{conflict} is present. Similarly, if a triple (\textbf{entity1}, \textit{country}, \textbf{entity2}) is present and \textbf{entity1} is of type \textit{location} or \textit{organization}, (\textbf{entity1}, \textit{located in}, \textbf{entity2}) is considered to be present as well. We show the list of co-occurring relations in Table~\ref{tab:cooccur-rel}. 

The logical rules automatically add relation triples to the Re-DocRED dataset, but it is possible that erroneous triples are also added due to corner cases. However, according to the human evaluation described in Appendix~\ref{app:comparisonwithscratch}, the precision of our Re-DocRED dataset is still higher than \citep{Huang2022DoesRP}.

\begin{table}[ht]
\centering
\begin{tabular}{lc} 
\toprule
Relation & \multicolumn{1}{l}{Co-occurring relation}  \\ 
\midrule
country  & located in                                 \\
conflict & participant of                             \\
\bottomrule
\end{tabular}
\caption{List of co-occurring relations.}
\label{tab:cooccur-rel}
\end{table}


\begin{table}[ht!]
\centering
\begin{tabular}{ll} 
\toprule
Relation            & Inverse relation         \\ 
\midrule
author              & notable work             \\
performer           & notable work             \\
producer            & notable work             \\
composer            & notable work             \\
director            & notable work             \\
lyrics by           & notable work             \\
participant         & participant of           \\
participant of      & participant              \\
has part            & part of                  \\
sibling             & sibling                  \\
series              & has part                 \\
spouse              & spouse                   \\
characters          & present in work          \\
conflict            & participant              \\
parent organization & subsidiary               \\
subsidiary          & parent organization      \\
follows             & followed by              \\
followed by         & follows                  \\
father              & child                    \\
replaced by         & replaces                 \\
head of government  & applies to jurisdiction  \\
replaces            & replaced by              \\
legislative body    & applies to jurisdiction  \\
head of state       & applies to jurisdiction  \\
mother              & child                    \\
part of             & has part                 \\
sister city         & sister city              \\
capital             & capital of               \\
\bottomrule
\end{tabular}
\caption{List of inverse relations.}
\label{tab:inverse-rel}
\end{table}

\end{document}